\documentclass[letterpaper, 10 pt, conference]{ieeeconf}  

\IEEEoverridecommandlockouts                              

\usepackage{times}
\usepackage{epsfig}
\usepackage{graphicx}
\usepackage{amsmath}
\usepackage{amsfonts} 
\usepackage{hyperref}
\usepackage{cite}

\usepackage{algorithm}
\usepackage[noend]{algpseudocode}

\usepackage{subcaption}
\DeclareCaptionSubType*[Alph]{table}
\DeclareCaptionLabelFormat{mystyle}{Table~\bothIfFirst{#1}{ ̃}#2}
\usepackage{multirow}
\usepackage{array}
\newcolumntype{P}[1]{>{\centering\arraybackslash}p{#1}}

\usepackage{xcolor}
\usepackage{comment}

\title{\LARGE \bf
A Personalized Household Assistive Robot that Learns and Creates New Breakfast Options through Human-Robot Interaction
}

\author{Ali Ayub$^{1*}$, 
Chrystopher L.\ Nehaniv$^{1}$,  and Kerstin~Dautenhahn$^{1}$
\thanks{This research was undertaken, in part, thanks to funding from the Canada 150 Research Chairs Program and thanks to an International Research and Partnership Grant at University of Waterloo.}
\thanks{$^{1}$University of Waterloo, Waterloo, Ontario N2L 3G1, Canada}
\thanks{{\tt\small $\{$*\,a9ayub, cnehaniv, kdautenh$\}$@uwaterloo.ca}}
}

\begin{document}

\maketitle
\thispagestyle{empty}
\pagestyle{empty}

\begin{abstract}
\label{sec:Abstract}
For robots to assist users with household tasks, they must first learn about the tasks from the users. Further, performing the same task every day, in the same way, can become boring for the robot's user(s), therefore, assistive robots must find creative ways to perform tasks in the household. In this paper, we present a cognitive architecture for a household assistive robot that can learn personalized breakfast options from its users and then use the learned knowledge to set up a table for breakfast. The architecture can also use the learned knowledge to create new breakfast options over a longer period of time. The proposed cognitive architecture combines state-of-the-art perceptual learning algorithms, computational implementation of cognitive models of memory encoding and learning, a task planner for picking and placing objects in the household, a graphical user interface (GUI) to interact with the user and a novel approach for creating new breakfast options using the learned knowledge. The architecture is integrated with the Fetch mobile manipulator robot and validated, as a proof-of-concept system evaluation in a large indoor environment with multiple kitchen objects. Experimental results demonstrate the effectiveness of our architecture to learn personalized breakfast options from the user and generate new breakfast options never learned by the robot.
\end{abstract}

\section{Introduction}
\label{sec:introduction}
\noindent
With a rapid increase in the aging population worldwide \cite{iriondo18,fuss20}, research is being conducted to develop autonomous robots that can assist older adults in their homes. These assistive robots are being designed for various roles, such as caretakers, cleaning robots, and home assistants \cite{Matari17,ayub2022don,petrecca_how_2018,shah2023my}. To create robots that can assist users with household tasks, the robots will first need to learn the preferences of the users related to the assistive tasks. For example, for the task of setting up a table for breakfast, the robot must first learn the different kinds of breakfasts that the user likes. Further, after learning the user preferences, the robot must find creative ways to perform the assistive tasks, because performing the same task every day can become boring for the user. For example, setting up the same breakfast option for the user over multiple days could become boring and the user might want to try new things. Therefore, in this paper, our goal is to develop a computational architecture that can allow a household assistive robot to learn different breakfast options from its user, use the learned knowledge to set up a table for breakfast, and also create new breakfast options for the user.

For a household assistive robot to perform tasks, it needs the semantic knowledge of the household i.e. objects (e.g. bowl, spoon) and related contexts (e.g. kitchen). The robot must also be able to reason on the semantic knowledge to perform tasks using the objects in the household. Extensive research has been conducted in recent years to create semantic reasoning architectures for performing assistive tasks in household environments \cite{kazhoyan2021robot,liu2022service}. Most of these works use a pre-specified knowledge base to perform household tasks. However, in the real world, different users can have different preferences about the tasks that they need assistance with. Therefore, for such cases, we need to develop personalized household robots \cite{Dautenhahn04} that can learn about the tasks that the users need assistance with, from the users. Research has also been conducted on creativity for robots. Most research in this field has been on developing cognitive architectures for social robots to create new artistic drawings \cite{augello2016analyzing}, or for humanoid robots to perform creative dance moves~\cite{infantino2016robodanza,manfre2016exploiting}. However, these works are not directly applicable to household assistive robots for completing tasks in creative ways.

\begin{figure*}
\centering
\includegraphics[width=0.7\linewidth]{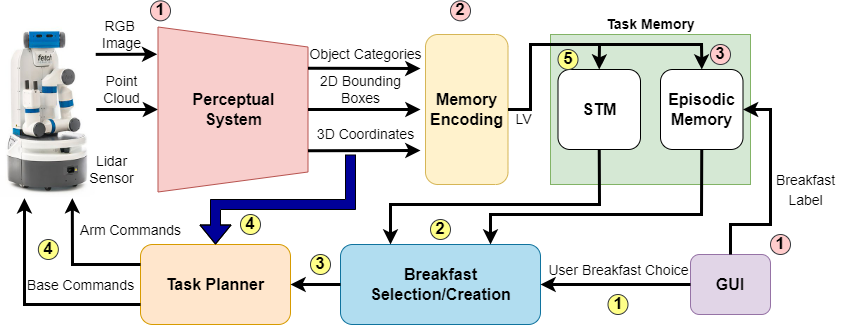}
\caption{\small Our complete architecture for learning and setting up breakfast options in a household. Sensory inputs from the Fetch robot are processed through the perceptual system and encoded into latent variables, which are stored in the episodic memory during the learning phase, and stored in STM to track breakfasts eaten by the user over multiple days. The breakfast creation module can use the data in the episodic memory to create new breakfast options. The task planner can plan lower-level commands for the robot's actuators to set up a table for breakfast. The wide, dark blue line indicates that all three outputs from the perceptual system are passed on to the task planner. Circled numbers show the flow of information in the architecture, with pink-colored numbers for the learning process and yellow-colored numbers for the breakfast setup. Processes that run in parallel are tagged with the same number.}
\label{fig:architecture_grocery}
\end{figure*}

In this paper, we develop a cognitive architecture that allows a robot to learn different breakfast options using the objects in the household from its user, set up the learned breakfast options on a table upon request from the user, and create new breakfast options for the user over the long term. The architecture allows the robot to interact with its user using a graphical user interface (GUI) and learn different breakfast options. Inspired by the dual memory theory of mammalian memory \cite{mcclelland95}, the breakfast options taught by the user, grounded in the processed sensory data of the robot, are stored in the long-term episodic memory. The architecture also keeps track of different breakfasts eaten by the user over multiple days and stores them in short-term memory (STM). The architecture can access the learned knowledge from the episodic memory and plan lower-level actuator commands for the robot to set up a table for the learned breakfasts. The architecture can further reason on the knowledge stored in the episodic memory to generate a semantic knowledge graph which can be used to create new breakfast options. The user can ask the robot to set up a previously learned breakfast or create a new breakfast option through the GUI. We integrate the proposed architecture on the Fetch mobile manipulator robot \cite{Wise16} and test it in a large indoor space with 9 common kitchen objects. Experimental results confirm that the robot can accurately learn different breakfast options from the user and set them up on a table. The results also show that the robot can create various new breakfast options that were never observed by the robot in its experience in the household context.

\section{Related Work}
\label{sec:related_work}
\noindent  
Socially assistive robots have been developed in recent years that can be interactive meal partners for older adults in long-term care homes \cite{McColl14,mccoll2013meal}. These robots, however, only interact with older adults to suggest different meal options and do not physically perform the task of setting up the table for a meal. Various cognitive architectures have been developed that can use the semantic knowledge of a household environment and physically perform tasks in the household, such as fetching an object, setting up a table for breakfast, cleaning a table \cite{kazhoyan2021robot,liu2022service,wang2020home}. Although these robots can perform different tasks in a household environment, they perform only a pre-programmed set of tasks, and they do not adapt to the preferences of their users. For example, the mobile manipulator robot in \cite{kazhoyan2021robot} can set up a table for only one type of breakfast. This can also get boring for the users if the robot sets up the same breakfast every single day over multiple weeks. In such cases, the robot must create new breakfast options for its users.

Research for developing creative robots has been limited to creating artistic drawings or dancing robots. For example, Augello et al. \cite{augello2016analyzing} develop a cognitive architecture for social robots that can create a new drawing while collaborating with a human. Infantino et al. \cite{infantino2016robodanza} and Manfre et al. \cite{manfre2016exploiting} develop cognitive architectures to enable creativity in humanoid robots so that they can dance in pleasant manners. These works, however, are not applicable to household assistive robots that can perform household tasks in creative ways. Research has also been conducted on developing cognitive architectures that can allow social robots to stimulate creativity in children \cite{alves2020software,elgarf2021once}. These architectures, however, do not allow a robot to be creative but rather stimulate creativity in children. 

With the advent of deep learning, generative adversarial networks (GANs) have been developed that can generate new data the model never learned \cite{zhang2021decorating,mateja2021towards,Ayub_iclr20}. These networks can learn general semantic representations about different household contexts (e.g. bedroom) from a large amount of training data, and then generate new images that were never seen by the model. One of the main limitations of these models is that they can generate many random images which do not belong to any context, such as creating random images that do not look like a bedroom context. Therefore, they cannot be applied to make assistive robots creative, as the robot would make many mistakes, which can hurt the trust of its user towards the robot \cite{robinette2017effect}. Further, GANs also require a large amount of training data to learn, which might be infeasible in real-world situations where the robot learns from the supervision provided by its users. Real users (especially older adults) would be unwilling to provide hundreds and thousands of examples of a single task to teach the robot. In this paper, we use Gaussian processes \cite{miller1966probability} as generative models to create new breakfast options, as these models have been shown to work with limited data \cite{ayub2020allowed}.

\section{Contextual Memory System for a Creative Robot}
\label{sec:methodology}
\noindent Figure \ref{fig:architecture_grocery} shows our cognitive architecture for a creative breakfast setting robot. Different computational modules in the architecture were integrated using ROS on the Fetch mobile manipulator robot. Note that all the modules are stand-alone, therefore they can be reused as blocks in different frameworks. These modules are described below: 

\subsection{Robot's Sensors}
\noindent The Fetch mobile manipulator robot was used for this project \cite{Wise16}. Fetch consists of a mobile base and a 7 DOF arm. The robot also contains an RGB camera, a depth sensor and a Lidar sensor. These sensors can be used for 3D perception, slam mapping, and obstacle detection in the robot's environment. In our architecture, the mobile base, the 7 DOF arm, and all three sensors are used for perception, manipulation, mapping, and navigation in an indoor environment. 

\subsection{Perceptual System}
\label{sec:perceptual_system}
\noindent The perceptual system of the architecture takes an RGB image and point cloud data as input from the robot's sensors, and parses this data into separate objects. We use the YOLOv2 object detector \cite{Redmon_2016_CVPR} for the detection of objects in the RGB images. The 2D bounding boxes from YOLO are converted into 3D coordinates using the point cloud data. We collected $\sim$5000 images of 9 household objects used in our experiments and trained the YOLO object detector on the collected data. The perceptual system, thus, parses the input images and outputs the object categories, 2D bounding boxes and 3D coordinates for all the objects in the image.

\subsection{Memory Encoding}
\label{sec:memory_encoding}
\noindent
The data obtained from the robot's sensors or the perceptual system must be encoded into a low dimensional feature space (also called a \textit{latent variable}), before it can be used to reason about the entities in the world (e.g. objects in the household). In this paper, we encode the processed sensory inputs by the perceptual system, using \textit{conceptual spaces} \cite{gardenfors2004conceptual,Douven20}. In cognitive science, a Conceptual Space is a metric space in which entities are characterized by quality dimensions. Conceptual spaces have mostly been used in cognitive science for category learning, where the dimensions of a latent variable (LV) in a conceptual space represent the category features. In this paper, we use a conceptual space LV to represent different breakfast setups (such as \{cereal, milk, bowl, spoon\} make a breakfast setup), where the features of the LV represent the collection of objects in the breakfast setup represented by the LV. Further, as each breakfast setup contains food items such as cereal, milk, etc and utensils such as spoon, bowl, etc, we also encode this information about the objects in another LV. We term this LV, a food-context LV to differentiate it from the object LV for the breakfast options. This information can help the architecture generate creative breakfast setups (Section \ref{sec:reasoning_module}). 

\subsection{Short-Term Memory (STM)}
\label{sec:stcm}
\noindent Once an input image of a breakfast setup is encoded into a latent variable, it is stored in the short-term memory (STM) of the architecture. The size ($k$) of STM is set as a hyper-parameter to allow the architecture to store encoded images for a certain number of days. Once STM is full, data stored from earlier days is removed to make room for more data. 

STM tracks the breakfast eaten by the user over multiple days. Using the data stored in STM, the architecture can suggest new breakfast options that the user has not eaten in previous days. Formally, let's consider there are $n$ number of breakfast options stored in the episodic memory as LVs $X=\{x_1, x_2, ..., x_n\}$. Over the course of $k$ (hyperparameter in STM) number of days, the user eats different breakfast options, where $M=\{m_1, m_2,..., m_n\}$ represents the total number of times each of the $n$ breakfast options was eaten by the user. From this set, the robot can find the breakfast options that have been least eaten by the user over $k$ days as $\arg \min  M$ and set it up on the table. If multiple breakfast options were eaten the least number of times, then the robot randomly chooses one of these breakfast options.

\subsection{Episodic Memory}
\label{sec:episodic_memory}
\noindent The episodic memory stores different breakfast options taught by the robot's user. As different users can have different breakfast preferences, it is not possible to store a general set of breakfast options. Therefore, the robot must learn about these preferences by interacting with the user. 

In our architecture, a user can initiate a learning session using a GUI (details in Section \ref{sec:gui}) and provide examples of different breakfast setups. The robot captures the breakfast setups as images using its sensors. The perceptual system (Section \ref{sec:perceptual_system}) processes the training images which are then encoded into latent variables (Section \ref{sec:memory_encoding}). The encoded LVs (both object LVs and food-context LVs) are then stored in the episodic memory, which can be accessed later to set up a table for breakfast.

\subsection{Creating New Breakfast Options}
\label{sec:reasoning_module}
\begin{figure}[t]
\centering
\includegraphics[width=1.0\linewidth]{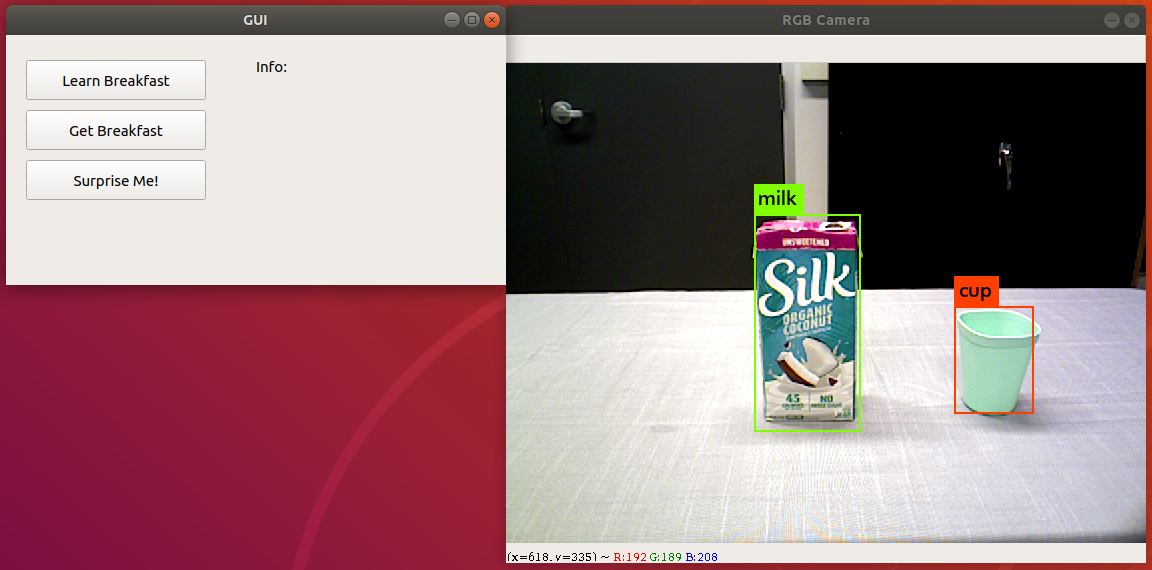}
\caption{\small The graphical user interface (GUI) used to interact with the robot. The head camera output of the Fetch robot with objects detected through YOLO is in the window to the right. The window to the left shows three buttons that can be used to teach the robot a new breakfast option, ask the robot to bring a previously learned breakfast option and ask the robot to create a new breakfast option.}
\label{fig:gui_grocery}
\end{figure}

\noindent The user can also ask the robot to surprise them (see Figure~\ref{fig:gui_grocery}) by creating a new breakfast option that the user never taught the robot i.e. such a breakfast option does not exist in the episodic memory. We define a creative breakfast as a new combination of food and utensil items that were never directly learned by the robot from the user. To achieve this, we use the object LVs stored in the episodic memory to find the mean $\mu$ and covariance matrix $\Sigma$ for a Gaussian distribution in a Gaussian process. We generate a pseudo-LV\footnote{The sampled LVs are termed as pseudo-LVs because they are not real LVs learned from the user.} after sampling the Gaussian distribution. However, the pseudo-LV can be the same as one of the object LVs stored in the episodic memory i.e. it is not a new breakfast option. Therefore, if the pseudo-LV is the same as any of the object LVs in the episodic memory, we continue to resample the Gaussian distribution until we get a pseudo-LV that is different from the object LVs in the episodic memory. 

The new pseudo-LV, however, can be an invalid breakfast setup. For example, \{cereal, milk, spoon\} is an invalid breakfast setup as it does not contain any container (such as a bowl) to pour cereal and milk. To fix such cases, we find the conditional relationships among various objects that are used in different breakfast setups stored as LVs in the episodic memory. Using these conditional relationships we infer logic-based rules to generate a knowledge graph, which can be used to fix invalid breakfast setups. 

\begin{figure}[t]
\centering
\includegraphics[width=1.0\linewidth]{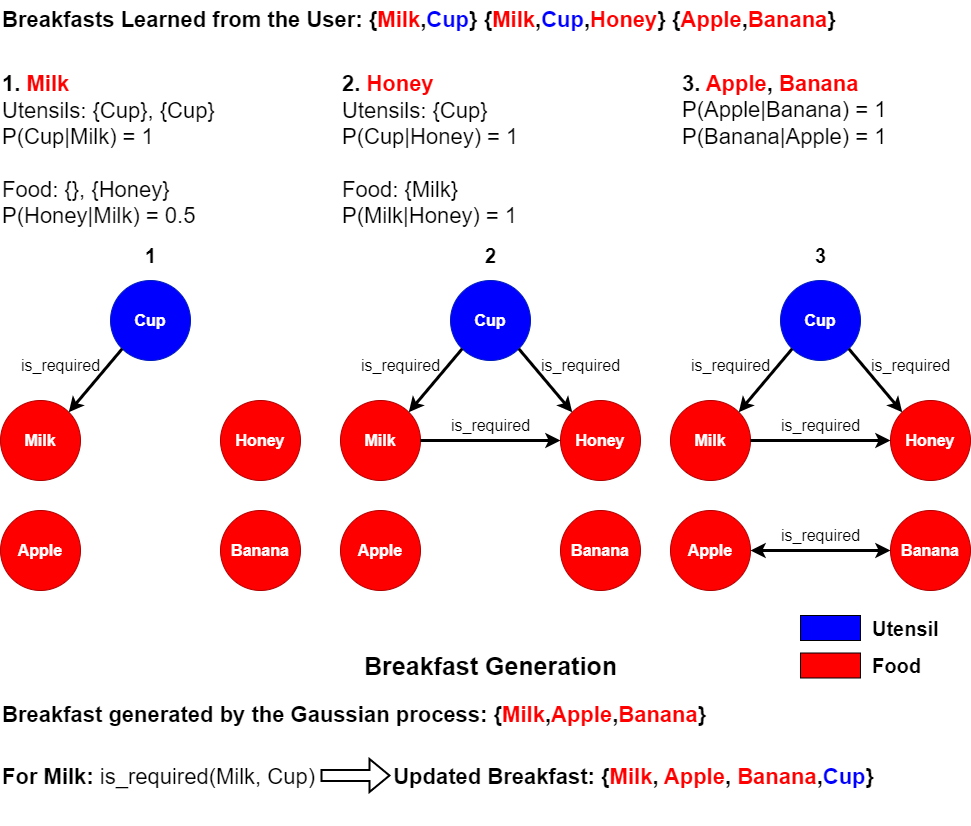}
\caption{\small An example of logic-based knowledge graph generation using the three breakfast options learned from a user. Digits next to visualizations correspond to the rule generation process for each food object. For \textit{Apple} and \textit{Banana}, we show the final probabilities only. Note that the logic-based rules are learned solely from the data shown by the user to the robot, therefore, some of the rules could be imperfect or unconventional, such as the requirement of \textit{Apple} and \textit{Banana} with each other. (Bottom) An example of fixing the generated breakfast option using the logic-based rules from the knowledge graph.}
\label{fig:creativity_example}
\end{figure}

We use the food-context LVs in the episodic memory to determine the dependency of different food items on a combination of other food items and utensils. To achieve this, let's consider $n$ LVs in the episodic memory, and consider a food object represented by dimension $i$ in the LVs. For each $i$th food item, we consider all the breakfast setups (say $r$ LVs) where this food item exists. Among the $r$ LVs, we first calculate the probability ${\rm{P}}(i|no\_utensil)$, i.e. if the $i$th food item does not require a utensil to be present in a breakfast setup. If ${\rm{P}}(i|no\_utensil)>0$, there is at least one breakfast setup where the food item is not accompanied by a utensil, therefore the food item can be a part of a breakfast setup without a utensil. Otherwise, if ${\rm{P}}(i|no\_utensil)=0$, the food item requires at least one utensil. In this case, we go through all $r$ LVs to find different combinations of utensils that the food item depends on, as a food item could depend on multiple utensils, e.g. cereal would depend on a spoon and a bowl. For this, let's consider that there are a total of $m$ utensils present in the $r$ LVs. For each $j$th utensil present in the $r$ LVs, we find the conditional probability ${\rm{P}}(j|l)$ with all the other $l=\{1,...,m\}$ utensils in the $r$ LVs. ${\rm{P}}(j|l)$ represents the probability that $j$th utensil exists given that $l$th utensil exists in the same LV. ${\rm{P}}(j|l)$ is determined as follows:

\begin{equation}
    {\rm{P}}(j|l) = \frac{\sum_{q=1}^r z_q^j \mbox{ such that   }z_q^jz_q^l>0}{\sum_{q=1}^r z_q^l \mbox{ such that }z_q^l>0 } ,
\end{equation}

\noindent where $z_q^l$ represents the value of $l$th utensil item in the $q$th food-context LV. If ${\rm{P}}(j|l)=1$, utensil $j$ must exist when utensil $l$ exists in an LV accompanied by the $i$th food item. As a result, we get an $m\times m$ matrix representing the dependency of utensils on other utensils that accompany the $i$th food item in $r$ LVs. Using this dependency matrix, we find all the utensil items that are independent of other utensils or that are interdependent with other utensils i.e. for two utensils $j$ and $l$, ${\rm{P}}(j|l)$=1 and ${\rm{P}}(l|j)$=1. The resulting set represents combinations of different utensils that the $i$th food item depends on. These sets are then used to generate a logic-based knowledge graph based on $is\_required$ relationships (see Figure \ref{fig:creativity_example} for an example). Note that the food item requires only one of the dependent utensil combinations to be present in the breakfast setup, not all the combinations. For example, milk must either be accompanied by a cup for drinking or \{bowl, spoon\} in a cereal breakfast. Figure \ref{fig:creativity_example} shows a simple example of generating a logic-based knowledge graph from the learned breakfast options in memory.  

After finding the dependencies on utensils, the same process is repeated to determine if a food item depends on other food items. After this process, we can find a combination of objects (foods or utensils) that each food item in the LVs depends on for valid breakfasts. We do not find a separate list of dependent objects for utensils as these items are only needed to accompany the food items in breakfast setups. Note that the knowledge graph is generated based on the breakfast options taught by the user, so the dependency rules encoded in the graph are personalized to the user. Experimental results in Section \ref{sec:experiments} confirm this. 

Using the logic-based knowledge graph, we can determine if a feature dimension in a pseudo-LV satisfies its dependency on other items. If a feature dimension is not accompanied by its dependent items, we manually add the dependent items in the pseudo-LV (see Figure \ref{fig:creativity_example} for an example). Finally, after the dependency check, the pseudo-LV is decoded using the inverse of the procedure in Section \ref{sec:memory_encoding} to get the objects in the new breakfast option. The object names/labels are then passed on to the task planner.

\subsection{Task Planner}
\label{sec:task_planner}
\noindent The task planner gets the decoded breakfast option from the creativity module (Section \ref{sec:reasoning_module}), and plans lower-level actions to be taken by the robot to set up a table for breakfast. The task planner passes lower-level commands to the mobile base and the arm of the robot to move and fetch objects from the kitchen to the dining table. 

\subsection{Graphical User Interface}
\label{sec:gui}
\noindent A simple graphical user interface (GUI) is integrated with the architecture to allow the robot to communicate with the user. The GUI allows the user to initiate a teaching session with the robot where the user can show the robot different breakfast setups on a table. The user physically places the set of objects in a breakfast setup on the table in front of the robot's camera (see Figure \ref{fig:gui_grocery}). The user can provide the name for the breakfast option by typing it in a textbox. The robot captures the breakfast data using the RGB camera and the depth sensor and then encodes and stores the breakfast option in the episodic memory (Section \ref{sec:episodic_memory}). 

The GUI also allows the user to ask the robot to set up a table for breakfast. The user can type in the name of the breakfast that they want, ask the robot to set up the table for breakfast without typing any particular breakfast name or ask the robot to surprise them by creating a new breakfast option. After getting the input from the user, the architecture can use a combination of all the modules to allow the robot (Section \ref{sec:task_planner}) to set up the table for breakfast.

\section{Experiments}
\label{sec:experiments}
\noindent In this section, we first describe the experimental setup and the implementation details. We then describe two experiments to evaluate the performance of our architecture for learning different breakfast options from the user, setting them up on the table, and creating new breakfast options. For all the experiments reported in this section, the experimenters take the role of a user.

 \begin{table}[t]
\centering
 \caption{\small Results of learning and setting up 7 breakfast options by the robot over 15 runs. Accuracy represents the ratio of the number of times a breakfast was correctly set up and the number of times it was chosen. STM shows the number of times a breakfast option was chosen by tracking data in short-term memory, without explicitly being asked by a user. LE represents the learning error.}
\begin{tabular}{ P{4.0cm}|P{1.2cm}|P{0.8cm}|P{0.8cm}}
     \hline
    \textbf{Breakfast Options} & \textbf{Accuracy} & \textbf{STM} & \textbf{LE} \\
     \hline
     milk, cup & 2/3 & 1 & 0 \\
     milk, cup, banana & 3/3 & 0 & 0\\ 
     milk, cereal, spoon, bowl & 1/2 & 1 & 0\\
     banana, milk, cereal, spoon, bowl & 1/1 & 0 & 0\\
     honey, milk, cereal, spoon, bowl & 1/2 & 0 & 0\\
     honey, milk, cup & 1/2 & 1 & 0\\
     apple, orange, banana & 1/2 & 0 & 0\\
    \hline
 \end{tabular} 
  \label{tab:known_breakfasts}
 \end{table}

\subsection{Experimental Setup}
\label{sec:experimental_setup}
\noindent We use the Fetch robot \cite{Wise16} and its associated ROS packages for all the experiments. We performed experiments in a large indoor space where we set up the kitchen and the dining area with realistic household objects. The indoor space is mapped using the Lidar sensor on the Fetch robot and an existing SLAM algorithm available from Fetch Robotics. Navigation in the environment was achieved using ROS packages provided by Fetch Robotics. Common household items/objects belonging to 9 categories (see Table \ref{tab:object_errors} for a list of graspable objects) are placed on three tables in the kitchen. Out of the 9 objects, 3 (\textit{Banana}, \textit{Bowl}, and \textit{Spoon}) were not graspable by the robot. Therefore, for breakfast setups that required these 3 objects, the user had to fetch the objects themselves. Manipulation of objects (pick and place) was achieved using ROS packages for gripper, arm, and torso control provided by Fetch Robotics.
The RGB camera and depth sensors on the Fetch robot were used for visual sensing of the environment. RGB images from the camera are passed through the perception module of the architecture which uses YOLOv2 \cite{Redmon_2016_CVPR} to detect and localize objects in the images (see Section \ref{sec:perceptual_system} for details). 

For all the experiments (unless mentioned otherwise), the user (experimenter) first teaches the Fetch robot different breakfast options on the dining table using the GUI (Section \ref{sec:gui}). The robot learns the breakfast options and stores them in episodic memory. As the user would eat breakfast once every day, we can `simulate' multiple days by asking the robot to set up a table for breakfast multiple times a day. For the short-term memory (STM) in the architecture, we set the hyper-parameter $k$ to 5 days. Examples of teaching breakfast options to the robot and testing the robot to set up known and new breakfast options are shown in the supplementary video.

\subsection{Experiment 1: Setting Up Known Breakfast Options}
\begin{figure}[t]
\centering
\includegraphics[width=1.0\linewidth]{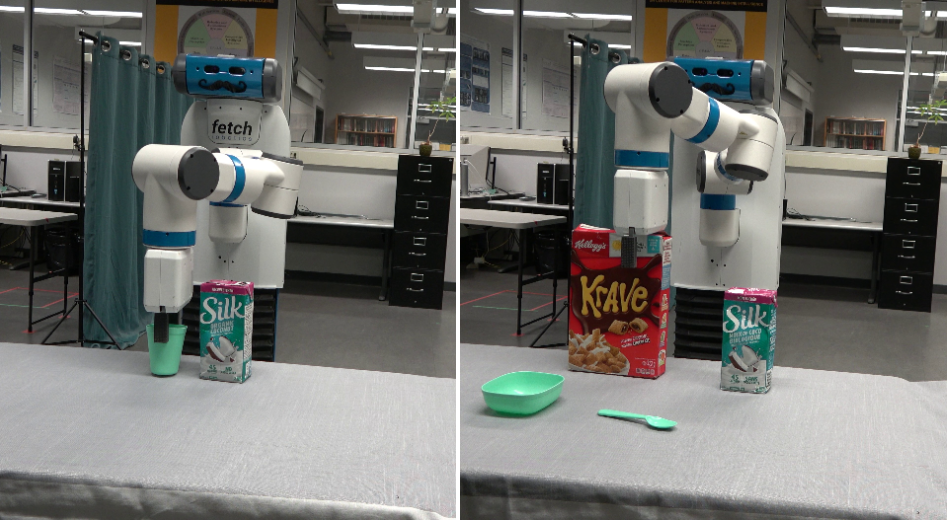}
\caption{\small Examples of two different breakfast options learned and set up by the Fetch robot.}
\label{fig:breakfast_options}
\end{figure}

\noindent In this experiment, we tested if the robot can learn breakfast options from the user and then set up the learned breakfast options when asked by the user. We taught the robot 7 different breakfast options as shown in Table \ref{tab:known_breakfasts}. Figure \ref{fig:breakfast_options} shows examples of 2 out of 7 breakfast options learned by the robot. The robot was then asked to set up a table for breakfast 15 times. In each of the 15 turns (except for the 5th, 10th, and 15th run), the user typed the name of the breakfast in the GUI to ask the robot to set up the table for a particular breakfast. We randomly chose a breakfast option to be typed in each turn. For the 5th, 10th, and 15th turn, the user did not type any breakfast name, therefore the robot used the data stored in STM to choose the least eaten breakfast to set up. For each breakfast setup, the robot moved all the graspable objects in the breakfast setup from the kitchen to the dining table. On average, it took the robot $\sim$4 minutes to set up a breakfast option on the table. 
 
 \begin{table}[t]
\centering
\caption{\small Results of setting up 7 breakfast options in 15 runs in terms of perceptual errors (PE), manipulation errors (ME), and grasping errors (GE) for graspable objects. Each column represents the ratio between the number of errors for manipulation of an object and the total number of times the object occurred in 15 runs.}
\begin{tabular}{ P{1.5cm}|P{0.7cm}|P{0.7cm}|P{0.7cm}  }
     \hline
    \textbf{Object} & \textbf{PE} & \textbf{ME} & \textbf{GE} \\
     \hline
     milk & 1/12 & 1/12 & 0 \\
     cup & 2/8 & 2/8 & 0 \\ 
     cereal& 0 & 1/5 & 0 \\
     apple & 0 & 0 & 0 \\
     orange & 1/2 & 0 & 1/2 \\
     honey& 1/4 & 2/4 & 0 \\
    \hline
 \end{tabular} 
  \label{tab:object_errors}
 \end{table}

Table \ref{tab:known_breakfasts} shows the results of setting up 7 breakfast options learned by the robot. All the breakfast options were learned correctly by the robot, and there was no learning error. The robot was able to correctly set up breakfast options in 10 out of 15 runs. As each breakfast setup required multiple objects, failing to fetch even a single object would result in an incorrect breakfast setup. Most of the breakfast setup failures happened because of a single object in the breakfast setup (more details below). There were three runs when the robot was asked to set up a breakfast option using STM. The robot correctly chose one of the least-eaten breakfast options in all three runs.

Table \ref{tab:object_errors} shows the results of the experiment in 15 runs, with three different kinds of errors for each graspable object used in the 7 breakfast setups. The most common error type was the manipulation error (ME), which occurred because of two reasons: (1) the motion planner could not find a path to reach the goal, (2) the perceptual system provided an incorrect pose estimate to pick the object (perceptual error (PE)). There were no object detection failures during the experiment because even if the robot failed to detect an object, it moved its head up and down until it found the correct object. Therefore, all of the perceptual errors happened during the 3D pose estimation of objects. Finally, there was only one grasping error for \textit{Orange}. The robot's arm was not low enough and because the orange is round it was not captured in the gripper of the robot. Objects with sharper edges, such as \textit{Milk} did not face this issue. These results confirm that our architecture can allow a robot to learn most breakfast options from the user and set up the learned breakfast options on a table.  

\subsection{Experiment 2: Creating New Breakfast Options}
\subsubsection{Experiment with a Robot}
\noindent In this experiment, we tested the ability of our architecture to allow a robot to create new breakfast options that were never learned by the robot. The experimental setup was the same as in experiment 1. The robot was started with the same 7 breakfast options in the beginning as in experiment 1. After that, we tested the robot 5 times to create and set up new breakfast options.

Table \ref{tab:creative_breakfasts} shows the five new breakfast options created by the robot. All five breakfast setups were valid setups because each food object was accompanied by the correct set of utensils. Two out of five breakfast options generated by the Gaussian process were invalid. For example, breakfast option 2 had bowl missing, and breakfast option 5 had cup missing. However, these objects were added by the architecture using the logic-based rules encoded in the knowledge graph for the food items (Section \ref{sec:reasoning_module}). Finally, note that there was no learning error (LE) encountered for these breakfast options because they were not learned from any example provided by the user to the robot. The values for perceptual and manipulation errors were consistent with the previous experiments. 

 \begin{table}[t]
\centering
\begin{tabular}{ P{4.5cm}|P{0.7cm}|P{0.4cm}|P{0.4cm}|P{0.4cm} }
     \hline
    \textbf{Breakfast Options} & \textbf{Valid?} & \textbf{LE} & \textbf{PE} & \textbf{ME} \\
     \hline
     milk, banana, honey, cup & Yes & 0 & 1/1 & 1/1 \\
     apple, milk, cereal, spoon, bowl & Yes & 0 & 0 & 1/1\\
     apple, honey, milk, cereal, spoon, bowl & Yes & 0 & 1/1 & 1/1\\
     milk, cereal, bowl, cup, spoon & Yes & 0 & 0 & 0\\
     apple, milk, banana, orange, cup & Yes & 0 & 0 & 0\\
    \hline
 \end{tabular}
  \caption{\small Five breakfast options created by the robot that were not taught by the user. LE, PE, and ME represent the learning error, perceptual error, and manipulation error, respectively.} 
  \label{tab:creative_breakfasts}
 \end{table}

\subsubsection{Simulated Experiments}
To further evaluate the effectiveness of our breakfast creativity algorithm, we tested the architecture in simulation to create 50 breakfast options. Note that in this case the architecture was only asked to suggest the breakfast option and the robot did not physically set up the generated breakfast option on the table. Out of the 50 breakfast options, 27 were the same setups as the ones stored in the episodic memory and were thus discarded. Out of the other 23 options, 7 were invalid options generated by the Gaussian process. However, these invalid options were corrected using the logic-based knowledge graph for the food items. Overall, out of the 23 new breakfast options, 6 were duplicates, so there were 17 distinct new options. These results confirm that our architecture can allow the robot to create and set up new breakfast options that were not learned by the robot. Further, our architecture was able to create more than double the breakfast options (17) it had learned by interacting with the user (7). However, the robot cannot generate a significantly large number of distinct breakfast options when learning from a few examples. 

We further test our approach on a \underline{larger scale} with a total of 25 objects, an initial set of 20 breakfasts and ask the creativity module to generate 200 breakfast options. Out of the 200 generated breakfasts, 65 were the same as the ones stored in the episodic memory and were therefore discarded. For the rest of the 135 breakfasts, 113 were invalid options but they were corrected by the logic-based knowledge graph for the food items. Finally, out of the 113 new breakfast options 36 were duplicates. Therefore, the architecture was able to generate 99 distinct new breakfast options from only 20 initial breakfast setups. These results confirm the scalability of our approach to larger datasets learned over the long term.

Finally, we tested our approach with some \underline{unconventional breakfast setups}. For example, we added a breakfast setup \{cereal, bowl\} as some users might eat cereal without any milk. Other examples of unconventional setups were \{peanut\_butter, bowl, spoon\}, \{yogurt, spoon\}, etc.. For this experiment, we had a total of same 25 objects as in the previous experiment, and 12 breakfast setups where 6 of the breakfast setups were unconventional. The creativity module generated 50 breakfast options, with 25 out of 50 being distinct new options. Interestingly, we noticed that the creative breakfast setups followed the dependency of food items learned through the data of the initial breakfasts. For example, the creativity module generated setups such as, \{apple, cereal, bowl, spoon, yogurt, peanut\_butter\} where cereal is not accompanied by milk. These results confirmed the ability of our architecture to personalize to their users' preferences even when creating new breakfast options. These results also show the effectiveness of our unique combination of data-driven learning, logic-based reasoning, and human-robot interaction.

\section{Conclusions}
\label{sec:conclusion}
\noindent This paper has presented an architecture for learning and setting up different breakfast options for the user. The architecture can also create new breakfast options that were never taught by the user. Extensive proof-of-concept system evaluations on a Fetch mobile manipulator robot demonstrate the ability of our architecture to allow a robot to accurately learn multiple breakfast options from the user and then set them up on a table upon request. The results also confirm the ability of the architecture to be able to track previously eaten breakfasts by the user to suggest new breakfasts, and even create multiple breakfast options that were never learned by the robot. We hope that this work will lead to designing more effective personalized household robots that can interact with, learn and provide long-term assistance to older adults in their own homes to support independent living.

\section{Limitations and Future Work}
\label{sec:limitations}
\noindent For all the experimental evaluations, the experimenter performed the role of the user. In the future, we hope to conduct a user study with real participants to investigate the usability of the system in real-world household environments. Further, the robot showed promising results with the chosen hyperparameter value for $k$ in the STM. However, we hope to perform more experiments in the future to analyze the effect of this hyperparameter on the choice of breakfast options. 

There were some objects that the robot struggled with when setting up different breakfasts, particularly because of the 3D pose estimation of objects. However, designing robust pose estimation and manipulation algorithms for complex household objects was out of the scope of this work. In the future, we hope to explore these limitations to scale up our approach to more realistic  household environments.

\addtolength{\textheight}{-12cm}   






{\small
\bibliographystyle{IEEEtran.bst}
\bibliography{main}
}

\end{document}